\documentclass[runningheads]{llncs}
\makeatletter
\newcommand{\printfnsymbol}[1]{%
  \textsuperscript{\@fnsymbol{#1}}%
}
\makeatother
\usepackage{graphicx}
\usepackage{array}
\usepackage{multirow}
\usepackage{graphicx}
\usepackage{comment}
\usepackage{amsmath,amssymb} 
\usepackage{color}
\newcolumntype{M}{>{\centering\arraybackslash}m{.2\textwidth}}
\newcolumntype{C}[1]{>{\centering\let\newline\\\arraybackslash\hspace{0pt}}p{#1}}
\newcolumntype{R}[1]{>{\raggedleft\let\newline\\\arraybackslash\hspace{0pt}}p{#1}}
\newcolumntype{L}[1]{>{\raggedright\let\newline\\\arraybackslash\hspace{0pt}}p{#1}}

\newcommand\Tstrut{\rule{-3pt}{2.6ex}}       
\newcommand\Bstrut{\rule[-0.9ex]{-3pt}{0pt}} 

\begin{document}
\title{SPNet: Multi-Shell Kernel Convolution for Point Cloud Semantic Segmentation} 
\titlerunning{SPNet}
%
 \author{Yuyan Li\thanks{Equal contribution} \and
 Chuanmao Fan\printfnsymbol{1} \and
 Xu Wang\printfnsymbol{1} \and
 Ye Duan}
 \authorrunning{Li, Fan. et al.}
%
 \institute{University of Missouri, USA \\
\email{\{yl235, cf7b6, xwf32, duanye\}@umsystem.edu}\\}
%
\maketitle              
\begin{abstract}


Feature encoding is essential for point cloud analysis. In this paper, we propose a novel point convolution operator named Shell Point Convolution (SPConv) for shape encoding and local context learning. Specifically, SPConv splits 3D neighborhood space into shells, aggregates local features on manually designed kernel points, and performs convolution on the shells. Moreover, SPConv incorporates a simple yet effective attention module that enhances local feature aggregation. Based upon SPConv, a deep neural network named SPNet is constructed to process large-scale point clouds. Poisson disk sampling and feature propagation are incorporated in SPNet for better efficiency and accuracy. 
We provided details of the shell design and conducted extensive experiments on challenging large-scale point cloud datasets. Experimental results show that SPConv is effective in local shape encoding, and our SPNet is able to achieve top-ranking performances in semantic segmentation tasks. 
\keywords{Point Cloud \and Semantic Segmentation \and Attention \and Deep Learning}

\end{abstract}
\section{Introduction}
Deep learning has achieved great success in image classification \cite{he2015delving,he2016deep,huang2017densely}, semantic segmentation \cite{long2015fully,yang2018denseaspp} and object detection \cite{girshick2014rich,girshick2015fast,he2017mask}. However, deep learning based point cloud analysis is still a challenging topic. One major reason is that point clouds are non-uniformly sampled from a large, continuous 3D space, making them lack of regular grid structure.
To tackle this, one straightforward approach is to voxelize point cloud into 3D regular grids and utilize standard 3D Convolutions \cite{tchapmi2017segcloud,le2018pointgrid}.  But the voxelization approach has a major limitation, the discretization step inevitably loses geometric information. 
To address this problem, many researchers have proposed approaches to directly process point clouds. One of the seminal works is PointNet proposed by Qi et al. \cite{qi2017pointnet}. It uses Multi Layer Perceptron (MLP) and global pooling to preserve permutation invariance and gather a combination of local and global feature presentation. This work is further improved in their follow-up work PointNet++ \cite{qi2017pointnet++}  which adds the local geometry pooling/sampling over local neighborhood.  Later works seek other ways to enhance local feature aggregation. For example, Pointweb \cite{zhao2019pointweb} constructs a dense fully-linked web, ShellNet \cite{zhang2019shellnet} conducts convolution based on statistics from concentric spherical shells. Besides point-based methods, several graph-based methods are proposed to capture 3D shape and structures of point clouds. Methods such as \cite{landrieu2018large,landrieu2019point} treat point clouds as nodes in a graph whose edges carry learnable affinitiy/similarity between adjacent points.

Recently, there has been another thread of research \cite{xu2018spidercnn,wang2018deep,wu2019pointconv,thomas2019kpconv,lei2019spherical} that proposes learnable kernel functions which define convolutional kernels on a continuous space. 
One of these approaches is KPConv \cite{thomas2019kpconv} introduced by Thomas et al. KPConv \cite{thomas2019kpconv} proposes kernel point operator that consists of a set of local point filters which simulate 2D image convolution processes. Features from unordered point clouds are aggregated on either rigid or deformable kernel points.
Using structural kernel points makes convolution feasible on a continuous space. 

\begin{figure}[t]
    \centering
    \includegraphics[width=12cm]{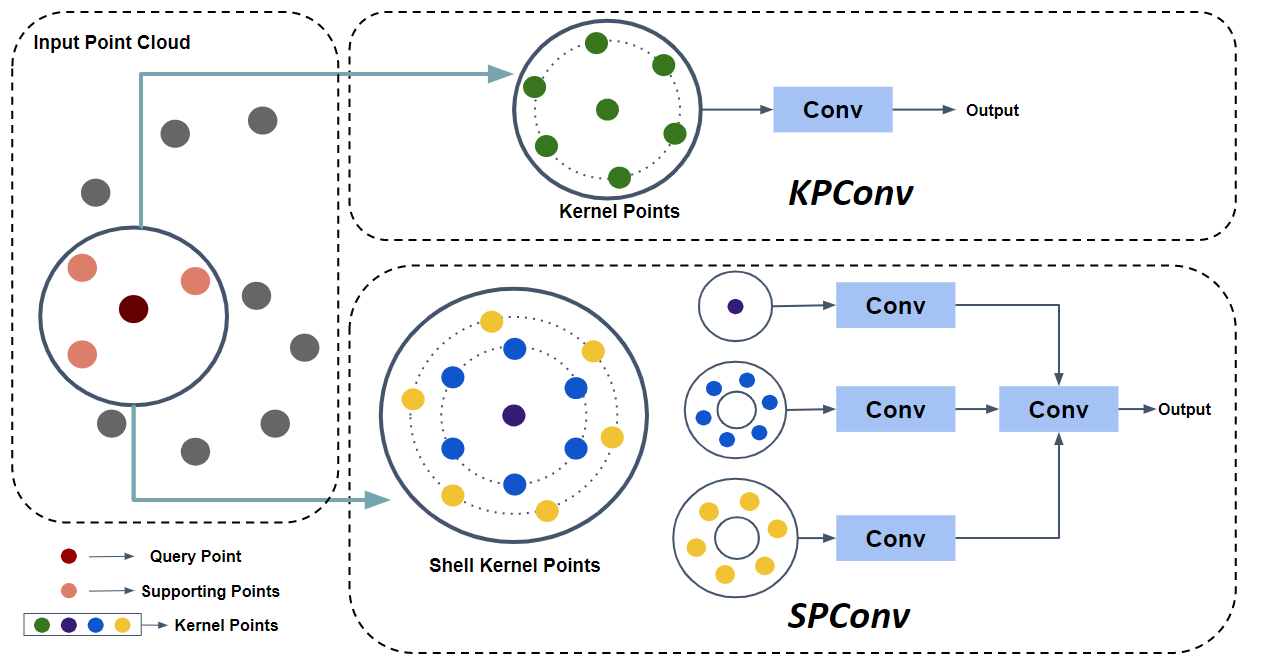}
    \caption{Comparison between SPConv and KPConv. For a query point, a range search is performed to locate supporting points. KPConv defines a set of kernel points to aggregate local features and performs point convolution. Our SPConv has multiple shells, each shell contains one set of kernel points. Point convolutions are conducted for shells individually to encode distinctive geometric information. An additional convolution layer is used to fuse shell outputs together, as an enhancement of structure correlation.}
    \label{fig:shell}
\end{figure}

Following this line of work, we propose a novel multi-shell kernel point convolution named SPConv. Our SPConv operator partitions local 3D space into shells, each shell contains a set of rigid kernel points which aggregate local supporting point features.  We perform kernel point convolution on each shell individually, then integrate the output features by an additional 1D convolution operation. The last convolution learns the contributions from shells and enhances shell correlation. An illustration of our SPConv is shown in Figure \ref{fig:shell}. Comparing to deformable KPConv \cite{thomas2019kpconv}, our method has an enhanced structure learning module, and does not require additional regularization during training. Furthermore, we find that incorporating low-level features such as color, normal, etc. in all layers for local feature re-weighting can be very effective for improving network performance. We propose two different approaches to accomplish the task, (1)Gaussian function based and (2)learning based. The first approach is hand-crafted and does not necessarily add GPU computation. The second approach has learnable weights and shows more robustness.

Using SPConv as our building block, we build a deep architecture SPNet. Similar to standard CNNs which utilize downsampling and upsampling strategy to reduce computation cost as well as to enlarge receptive field, we use Poisson disk sampling (PDS) for downsampling, and feature propagation (FP) for upsampling. 
In section 4, we evaluate the effectiveness of our methods on the most competitive indoor segmentation datasets. Notably, experimental results show that we achieve top-ranking performances.
Our main contribution is summarized as follows:
\begin{itemize}
  \item We propose a multi-shell kernel convolution operator that shows powerful local shape encoding ability.
  \item We introduce a simple yet effective attention mechanism for local neighbor feature re-weighting. This attention module improves performance and speeds up convergence.
  \item We present a comprehensive architecture design which outperforms stage-of-the-arts on challenging large-scale indoor datasets.  

\end{itemize}

\section{Related Work}

\textbf{View-based and Voxel-based Methods.}
One classic category of point cloud representations is multi-view representation. MVCNN \cite{su2015multi} renders 3D shape into images from various viewpoints and combines features from CNNs to predict point labels. However, these methods suffer from surface occlusion and density variation, which make it difficult to capture the internal structure of the shape. 
Another strategy is to convert point cloud into a 3D voxel structure which can be processed by standard 3D convolutions. VoxNet \cite{maturana2015voxnet} and subsequent work \cite{wu20153d,li2016fpnn} discretize point cloud into 3D volumetric grids. To improve efficiency on processing high resolution 3D voxels, recent researches  \cite{graham20183d,choy20194d} process volumetric data only on non-empty voxels.

\noindent\textbf{Point-wise MLP Methods.}
Point-based methods receive great attention since PointNet \cite{qi2017pointnet} was proposed. In PointNet \cite{qi2017pointnet}, points go through shared MLPs to obtain high dimensional features followed by a global max-pooling layer. In order to capture local neighborhood context, PointNet++ \cite{qi2017pointnet++} is developed by hierarchically applying pointnet in local regions. There are extensive works based on PointNet++. For example, PointWeb \cite{zhao2019pointweb} builds a dense fully connected web to explore local context, and uses an Adaptive Feature Adjustment module for feature refinement. ShellNet \cite{zhang2019shellnet} proposes a ShellConv operator with concentric spherical shells to capture representative features. 


\noindent\textbf{Point Convolution Methods.}
Some recent works define explicit kernels for point convolution. KCNet \cite{shen2018mining} develops a kernel correlation layer to compute affinities between each point’s K nearest neighbors and a predefined set of kernel points. Local features are acquired by graph pooling layers. SpiderCNN \cite{xu2018spidercnn} designs a family of Taylor polynomial kernels to aggregate neighbor features. PointCNN \cite{li2018pointcnn} introduces X-transformation to exploit the canonical order of points.  PCNN \cite{wang2018deep} builds a network using parametric continuous convolutional layers. SPH3D \cite{lei2019spherical} uses spherical harmonic kernels during convolution on quantized space to identify distinctive geometric features.
Our work is most related to KPConv \cite{thomas2019kpconv}, which defines rigid and deformable kernel points for feature aggregation. This convolution operator resolves point cloud ambiguity, alleviates varying density, and shows superior performances. Compared to KPConv \cite{thomas2019kpconv}, our SPConv enhances local structure correlation by incorporating shell-structured kernel points and learning on a larger neighborhood context.

\section{Methodology}

\subsection{Review on Kernel Point Convolution}
KPConv \cite{thomas2019kpconv} effectively resolves the point cloud ambiguity by placing manually designed kernel points in a local neighborhood. This convolution simulates image-based convolutions. A typical image based 2D convolution with a $(2m+1)\times (2m+1)$ kernel at location $i, j \in \mathbb{Z}$ is defined as:
\begin{equation}
    F * W = \sum_{x=-m}^{m}\sum_{y=-m}^{m}F(i-x, j-y) W(i,j)
\end{equation}
where $x, y \in \{-m, ...,m\}$, $W$ is the learnable weight, $F(i,j)$ is the feature for pixel $(i,j)$. Image based convolution describes a one-to-one relationship between single kernel and single image pixel. 
Similarly, for a 3D point $p\in \mathbb{R}^3$ with a local neighborhood of radius $R$, point convolution can be defined as:
\begin{equation}
    F * W = \sum_{k}^{K}F(p_k, p) W(k)
\label{eq:pointconv}
\end{equation}
where $F(p_k, p)$ is the aggregated features on kernel point $p_k$.  $p_k$ carries learnable matrix $W_k\in \mathbb{R}^{C_{in}\times C_{out}}$. $C_{in}$ and $C_{out}$ are input and output feature channels respectively. With proper aggregation approach, the structure of supporting points can be well captured and learned by weight $W$. There are two key components of this convolution, placements of kernel points and aggregation function. 

For a 3D point $x\in \mathbb{R}^3$ surrounded by neighboring points $x_j\in \mathbb{R}^3$ within a ball radius $R$, kernel points are distributed on the surface of a sphere with radius $r$, plus one point placed at center. Aggregated features $F(p_k, p)$ for kernel point $p_k$ are computed as the sum of the features carried by neighboring points that fall into the influenced radius $v$. These neighboring features are weighted based upon the Euclidean distance between $p_k$ and $p_j$. An illustration of KPConv is shown in Figure \ref{fig:shell}.
\begin{equation}
     F(p_k, p) = \sum_{p_j, \|p_j-p\|<R, \|p_k-p_j\|<v}F_{p_j} d(p_k, p_j)
\end{equation}
where $d(p_k, p_j)$ denotes the correlation of kernel point $p_k$ and a neighbor point $p_j$. This correlation can be calculated by a linear function:
\begin{equation}
     d(p_k, p_j) = max(0, 1-\frac{\|p_k-p_j\|}{v})
\end{equation}

\begin{figure}[tb]
    \centering
    \includegraphics[width=9cm]{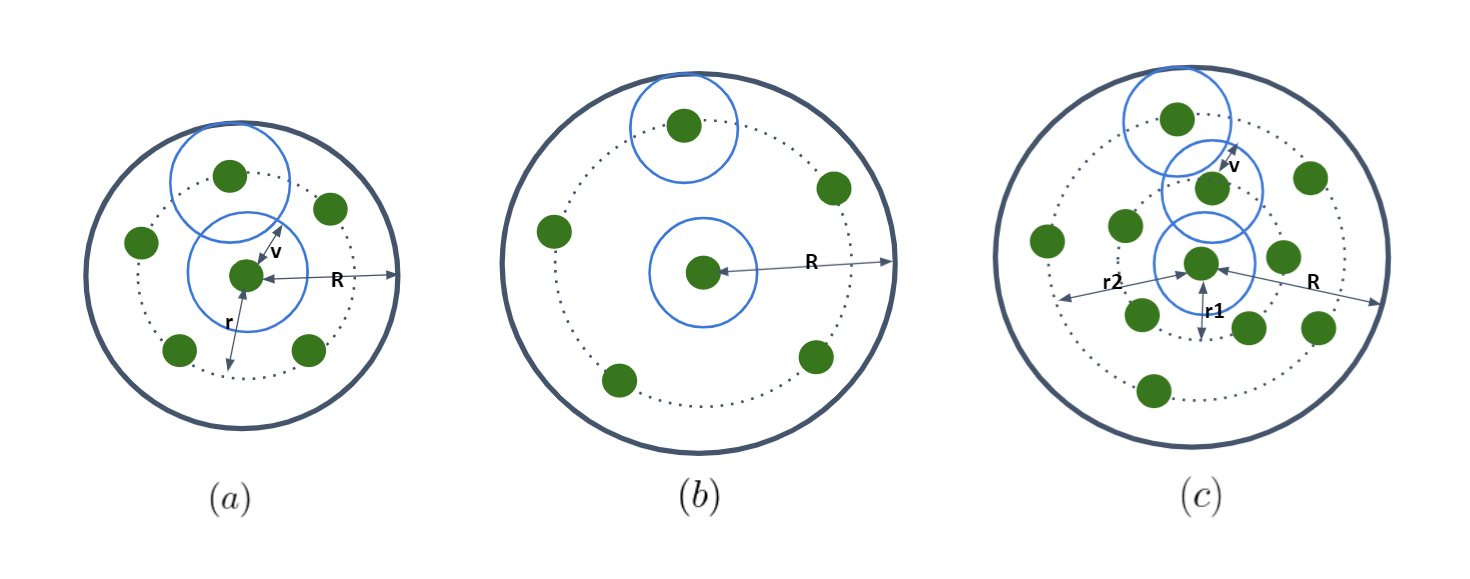} 
    \caption{A KPConv \cite{thomas2019kpconv} operator is defined in (a) with kernel point radius $r$, neighborhood size $R$, kernel influence radius $v$. Kernel points have overlapping influence regions. When enlarging neighborhood size $R$ to capture a larger context as shown in (b), kernel points become sparse and this may cause a loss of information for complex scenes with objects of different scales. Our SPConv (c) has multiple shells and keeps overlapping influence regions. $r_1$ and $r_2$ are kernel point radius for the second and third shell.}
    \label{fig:kernel}
\end{figure}

\subsection{SPConv}
We extend the work of KPConv and propose a new point convolution operator, SPConv. SPConv divides the local 3D space into a total of $N$ shells. Each shell has one set of kernel points. Specifically, the innermost shell contains one central kernel point $p_0$, for outter $n^{th}(n>1)$ shell, a set of kernel points $p_{1,m}, m\in M_n$ scatter on the surface of a sphere with radius $r_n$. Central kernel point impacts on a spherical region, and the $n^{th}$ shell forms a ring-shaped influence area. As a result, all kernel points cover a spherical space of radius $(r_N+v)$. We perform kernel point convolution on $N$ shells respectively, then stack shell features together along a new dimension. Finally, we use an additional convolution layer to further correlate shell structure.  
This convolution operator can be defined as follows:
\begin{equation}
    (F(x)*W_1)*W_2 = \sigma(\sum_n^{N}\sigma(F(x)*W_1) W_{2,n})
\end{equation}
where $\sigma$ refers to non-linear activation function. $W_1\in \mathbb{R}^{K\times C_{in}\times C_{out}/2}$ is the learnable weight matrix for kernel point convolution, $W_2\in \mathbb{R}^{N\times C_{out}/2\times C_{out}}$ is the weight matrix for shell correlation. To balance off efficiency and accuracy, we choose to use a total of 3 shells and 14 kernel points for the second and third shell respectively.  An illustration of our SPConv is shown in Figure \ref{fig:shell}.


A detailed illustration of the influence area of SPConv kernel points is shown in Figure \ref{fig:kernel}.  The central kernel point encodes features from points that are spatially close to the query point. Kernel points located far from the center tend to encode more contextual information. Therefore, we learn the features by shells based on the distance from kernel point to center such that the encoded features can be representative for each shell. Furthermore, the 1D convolution layer applies weight matrix on the fused shell features, which enhances structure learning across shells.
Our method aims to improve descriptive power of kernel points, so does KPConv \cite{thomas2019kpconv} deformable version. Although deformable kernels provide more flexibility, regularization imposed on offsets is mandatory to account for mis-shifts. By contrast, our kernels are rigid and do no require regularization. In section 4.4, we compare our evaluation results with deformable KPConv \cite{thomas2019kpconv}. Our method outperforms deformable KPConv with even less parameters.

\subsection{Feature Attention}
To further improve local feature encoding, we propose a feature attention module using low level features such as RGB or surface normal. 

We propose two approaches for local feature attention.
First approach is to apply a pre-defined Gaussian function:
\begin{equation}
    \omega_k = exp(-\frac{\|f(p)-f(s_k)\|}{2\sigma^2})
\end{equation}
where $\sigma$ is a parameter that needs to be manually set. 
The second approach is to use sequential MLPs:
\begin{equation}
    \omega_k = g(f(p)-f(s_k)), \; 
\end{equation}
 where $g$ is a sequence of MLPs activated by $ReLU$, except the last one which uses $sigmoid$. 
The final updated feature $f^{'}$ for point $s_k$ can be calculated as follows with a residual connection:
\begin{equation}
    f^{'}(s_k) =  \omega_k f(s_k) + f(s_k)
\end{equation}

 Both of the approaches improve performances and accelerate convergence speed. One issue for the first approach is that it is manually designed and not flexible. The second approach takes advantages of learnable weights and non-linear activations but adds more computation costs. 

\subsection{Network Architecture}
\begin{figure}[tb]
    \centering
    \includegraphics[width=13cm]{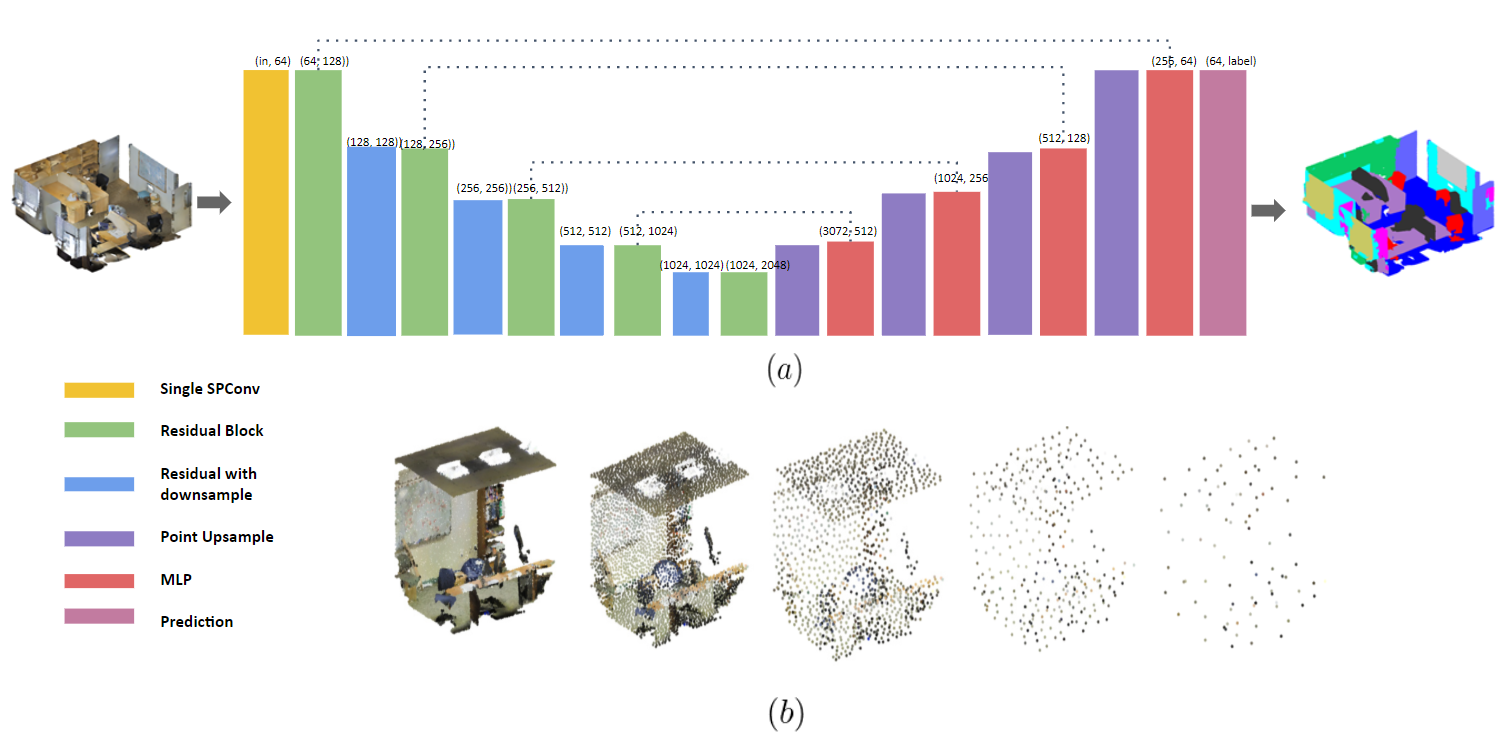}
    \caption{(a) Illustration of the network architecture. (b) Downsampling process by PDS at each level.}
    \label{fig:network}
\end{figure}

We build a deep encoder-decoding network with point downsampling and upsampling to accomplish semantic segmentation task. A detailed SPNet architecture is shown in Figure \ref{fig:network}(a).

\noindent\textbf{Downsampling Strategy}
Similar to works \cite{thomas2019kpconv,qi2017pointnet++}, we adopt downsampling to reduce computation load as well as to increase receptive field. 
In our work, we favors Poisson disk sampling (PDS) strategy to deal with the varying density. PDS controls spatial uniformity by Poisson disk radius $r_p$, thus maintaining a minimal distance between points. Unlike grid sampling as used in \cite{thomas2019kpconv} in which downsampled location are interpolated as the barycenter of a cell, PDS keeps the original locations of sub-sets and preserves shape patterns. Comparing to farthest point sampling (FPS) \cite{qi2017pointnet++}, PDS is faster when sampling large-scale points. A downsampling process by PDS is shown in Figure \ref{fig:network}(b).



\noindent\textbf{Upsampling Strategy}
With PDS, sampled points at each level are always a sub-set from input point sets. Therefore, we can accurately recover the point sampling patterns and gradually propagate features in decoder.  We adopt feature propagation module proposed in \cite{qi2017pointnet++}.
For a point $p_j$ at level $j$, its propagated features $f$ are calculated as:
\begin{equation}
    f = \sum_{k}^{K}w_k*f_k, \; 
    w_k = \frac{d_k^2}{\sum_{k}^{K}d_k^2}
\end{equation}
where $d_k$ is the inverse Euclidean Distance between $p_j$ and its $k_{th}$ nearest neighbor at level $j-1$. 

\section{Experiments}
In this section, we evaluate the performance of our network on large-scale semantic segmentation datasets. We provide extensive ablation studies to justify the effectiveness of our proposed methods. A comparison of scene segmentation results between existing methods and our is shown in Table \ref{tab:scene}.

\subsection{Datasets}
\textbf{Stanford Large-Scale 3D Indoor Spaces (S3DIS)}
The S3DIS dataset \cite{armeni2017joint} is a benchmark for large-scale indoor scene semantic segmentation. It consists of point clouds of six floors from three different buildings. Following the convention  \cite{qi2017pointnet,li2018pointcnn}, We perform experiments on both 6-fold and Area 5 to evaluate our framework.  For evaluation metrics, we use Overall point-wise accuracy (OA), and mean intersection over union (mIoU).
The detailed results for individual class are listed in Table \ref{tab:Table_S3DIS_area5} and Table \ref{tab:Table_S3DIS_6area}.  We can see that our method has the highest scores for several challenging classes, such as door, wall and board.

\noindent\textbf{Scannet}
The Scannet \cite{dai2017scannet} dataset contains more than 1500 scanned scenes annotated with 20 valid semantic classes. It provides a 1,201/312 data split for training and testing. The Scannet dataset is reconstructed from RGB-D scanner. We report
the per-voxel accuracy (OA) as evaluation metrics. As shown in Table \ref{tab:scene}, our framework achieves state-of-the-art performance.

\subsection{Implementation Details}
\textbf{Parameter Setting}
SPNet uses residual block similar to \cite{he2016deep}. Each block consists of one MLP for feature dimension reduction, one SPConv, and another MLP to increase feature dimension. SPNet consists of 5 encoding levels and 4 decoding levels, as shown in Figure \ref{fig:network}.
The kernel influence $v_0$ for the first encoding level is set to $0.04m$ for both S3DIS \cite{armeni2017joint} and ScanNet \cite{dai2017scannet}. For subsequent level $l$, kernel influence is increased to $v_l = 2^{l}v_0$. The rest of the parameters are adjusted according to $v_l$. For SPConv operator, we use a total of $K=3$ shells. Kernel points of the second and the third shell are initialized on the surfaces of spheres with radius $r_2 = 1.5v_l, r_3 = 3v_l$ respectively. Query neighborhood radius $R_l$ is set to $4v_l$, PDS radius is set to $0.75v$. For attention module, both color and normal are used to compute the attentional scores. For the Gaussian function, we set $\sigma=v_l$ at each level. 

\noindent\textbf{Network Training}
Our network is implemented using PyTorch \cite{paszke2017automatic} on a single Nvidia Titan RTX for all experiments. We use a batch size of 8, initial learning rate of 0.001. Optimization is done with Adam optimizer ($\beta_1=0.9, \beta_2=0.999$) \cite{kingma2014adam}. Learning rate decays by a factor of 0.3 every 50 epoch for S3DIS \cite{armeni2017joint}, and every 30 epoch for ScanNet \cite{dai2017scannet}.

\begin{table}[h]
\centering
\caption{Comparative 3D scene segmentation scores on S3DIS \cite{armeni2017joint}, ScanNet \cite{dai2017scannet} datasets. S3DIS \cite{armeni2017joint} scores are reported in metric of mean Intersection over Union(mIoU) including Area5 and 6-fold cross validation. ScanNet \cite{dai2017scannet} scores are reported as Overall Accuracy(OA) and mIoU. The symbol $'-'$ means the results are not available.}
    \begin{tabular}{ L{8em}  C{8em} C{8em} C{8em} }
    \hline
    Methods & S3DIS(mIoU) & S3DIS(mIoU) & ScanNet\\
     & Area5   & 6-fold   & (OA) \\
    \hline
    PointNet \cite{qi2017pointnet}   &  41.1  &  47.6 &  - \\
    PointNet++ \cite{qi2017pointnet++}   &  - &  54.5 & 84.5\\   
    DGCNN \cite{simonovsky2017dynamic}  &  - & 56.1   & -\\   
    SPGraph \cite{landrieu2018large}  & 58.0  &  62.1   & -\\ 
    ShellNet \cite{zhang2019shellnet} &  - &  66.8   & 85.2\\ 
    PointWeb \cite{zhao2019pointweb} &  60.3 &  66.7  & 85.9\\ 
    GACNet \cite{wang2019graph} & 62.9  &  - & - \\ 
    RandLA-Net \cite{hu2019randla} & - & 68.5 & -  \\
    SPH3D-GCN \cite{lei2019spherical} & 59.5 & 68.9  & - \\
    Point2Node \cite{han2019point2node} & 63.0  &  70.0  & 86.3\\ 
    KPConv(R) \cite{thomas2019kpconv} & 65.4  &  69.6  & -\\ 
    KPConv(D) \cite{thomas2019kpconv}  &  67.1 & 70.6   & -\\ 
    Minkowski \cite{choy20194d} & 65.4 & -  & - \\
    \hline
    \textbf{Ours} & \textbf{69.9}  & \textbf{73.7}  &  \textbf{89.5} \\ 
    \hline
    \end{tabular}

\label{tab:scene}
\end{table}

\begin{table*}[hbt!]
\setlength\tabcolsep{0.5pt}
\caption{Semantic segmentation mIoU and OA scores on S3DIS \cite{armeni2017joint} Area 5. }
\vspace{-0.5cm}
\begin{tiny}
\begin{center}
\begin{tabular}{L{1.7cm} | C{0.6cm} | C{0.6cm} | *{13}{C{0.7cm}}}
\hline
 Method	 & mIoU & OA & ceil.	 & floor	 & wall	 & beam	 & col.	 & wind.	 & door	 & chair	 & table	 & book.	 & sofa	 & board & clut.	\Bstrut\\
 \hline
PointNet \cite{qi2017pointnet}	& $41.1$	& $49.0$	& $88.8$	& $97.3$	& $69.8$	& $0.1$	& $3.9$	& $46.3$	& $10.8$	& $52.6$	& $58.9$	& $40.3$	& $5.9$	& $26.4$	& $33.2$	\Tstrut\\
 PointWeb \cite{zhao2019pointweb}	& $60.3$	& $87.0$	& $91.9$	& $\mathbf{98.5}$	& $79.4$	& $0.0$	& $21.1$	& $59.7$	& $34.8$	& $76.3$	& $88.3$	& $46.9$	& $69.3$	& $64.9$	& $52.5$	\\
 Point2Node	\cite{han2019point2node} & $62.9$	& $88.8$	& $93.8$	& $98.3$	& $83.3$	& $0.0$	& $35.6$	& $55.3$	& $58.8$	& $79.5$	& $84.7$	& $44.1$	& $71.1$	& $58.7$	& $55.2$	\\
KPConv(R) \cite{thomas2019kpconv} & $65.4$	& -	& $92.6$	& $97.3$	& $81.4$	& $0.0$	& $16.5$	& $54.5$	& $69.5$	& $90.1$	& $80.2$	& $74.6$	& $66.4$	& $63.7$	& $58.1$	\Tstrut\\
 KPConv(D) \cite{thomas2019kpconv}  & $67.1$	& - 	& $92.8$	& $97.3$	& $82.4$	& $0.0$	& $23.9$	& $58.0$	& $69.0$	& $\mathbf{91.0}$	& $\mathbf{81.5}$	& $\mathbf{75.3}$	& $75.4$	& $66.7$	& $58.9$	\Bstrut\\

 \hline
 \textbf{Ours}	& $\mathbf{69.9}$	& $\mathbf{90.3}$	& $\mathbf{94.5}$	& $98.3$	& $\mathbf{84.0}$	& $0.0$	& $24.0$	& $59.7$	& $\mathbf{79.8}$	& $89.6$	& $81.0$	& $75.2$	& $\mathbf{82.4}$	& $\mathbf{80.4}$	& $\mathbf{60.4}$	\\
 \hline
 \end{tabular}
 \end{center}
 \end{tiny}
 \label{tab:Table_S3DIS_area5} 
 \end{table*}

\begin{table*}[hbt!]
\setlength\tabcolsep{0.5pt}
\caption{Semantic segmentation mIoU and OA scores on S3DIS \cite{armeni2017joint} 6-fold. }
\vspace{-0.5cm}
\begin{tiny}
\begin{center}
\begin{tabular}{L{1.7cm} | C{0.6cm} | C{0.6cm} | *{13}{C{0.7cm}}}
\hline
Method	 & mIoU & OA & ceil.	 & floor	 & wall	 & beam	 & col.	 & wind.	 & door	 & chair	 & table	 & book.	 & sofa	 & board & clut.	\Bstrut\\
\hline
PointNet \cite{qi2017pointnet}	& $47.8$	& $78.5$	& $88.0$	& $88.7$	& $69.3$	& $42.4$	& $23.1$	& $47.5$	& $51.6$	& $54.1$	& $42.0$	& $9.6$	& $38.2$	& $29.4$	& $35.2$	\Tstrut\\
SPGraph	\cite{landrieu2018large} & $62.1$	& $85.5$	& $89.9$	& $95.1$	& $76.4$	& $62.8$	& $47.1$	& $55.3$	& $68.4$	& $69.2$	& $73.5$	& $45.9$	& $63.2$	& $8.7$	& $52.9$	\\
PointCNN \cite{li2018pointcnn} & $65.4$	& $88.1$	& $\mathbf{94.8}$	& $\mathbf{97.3}$	& $75.8$	& $63.3$	& $51.7$	& $58.4$	& $57.2$	& $69.1$	& $71.6$	& $61.2$	& $39.1$	& $52.2$	& $58.6$	\\
PointWeb \cite{zhao2019pointweb} & $66.7$	& $87.3$	& $93.5$	& $94.2$	& $80.8$	& $52.4$	& $41.3$	& $64.9$	& $68.1$	& $71.4$	& $67.1$	& $50.3$	& $62.7$	& $62.2$	& $58.5$	\\
KPConv(R) \cite{thomas2019kpconv} & $69.6$	& -	& $93.7$	& $92.0$	& $82.5$	& $62.5$	& $49.5$	& $65.7$	& $\mathbf{77.3}$	& $57.8$	& $64.0$	& $68.8$	& $71.7$	& $60.1$	& $59.6$ \\
Point2Node \cite{han2019point2node}	& $70.0$	& $89.0$	& $94.1$	& $97.3$	& $83.4$	& $62.7$	& $52.3$	& $72.3$	& $64.3$	& $75.8$	& $70.8$	& $65.7$	& $49.8$	& $60.3$	& $60.9$	\\
KPConv(D) \cite{thomas2019kpconv} & $70.6$	& -	& $93.6$	& $92.4$	& $83.1$	& $63.9$	& $54.3$	& $66.1$	& $76.6$	& $57.8$	& $64.0$	& $69.3$	& $\mathbf{74.9}$	& $61.3$	& $60.3$ \\
\hline
\textbf{Ours}	& $\mathbf{73.7}$	& $\mathbf{90.9}$	& $94.6$	& $\mathbf{97.3}$	& $\mathbf{85.0}$	& $45.2$	& $56.9$	& $\mathbf{82.1}$	& $63.4$	& $\mathbf{73.1}$	& $\mathbf{83.4}$	& $\mathbf{71.5}$	& $68.8$	& $\mathbf{68.6}$	& $\mathbf{67.8}$	\\
\hline
\end{tabular}
\end{center}
\end{tiny}
\label{tab:Table_S3DIS_6area} 
\end{table*}

\begin{table}[t]
\caption{Ablation studies evaluated on Area 5 of S3DIS \cite{armeni2017joint}.}
\vspace{-0.2cm}
    \centering
    \begin{footnotesize}
    \begin{tabular}{ L{20em}  C{4em} C{4em}}
    \hline
             &  mIoU & Gain$\Delta$\\
    \hline
    Baseline + grid sampling &  65.4  &  -\\
    \hline
    Baseline + grid sampling + FP &  65.4  &  -\\
    \hline
    Baseline + PDS  &  66.0  &  +0.6\\
    \hline
    Baseline + PDS + FP  &  67.7  &  +2.3\\
    \hline
    SPConv + PDS + FP  & 68.8  & +3.4 \\
    \hline
    Baseline + Attention + PDS + FP & 68.3 & +2.9 \\
    \hline
    SPConv + Attention + PDS + FP & 69.9 & +4.5 \\
    \hline
    \end{tabular}
    \end{footnotesize}
    \label{tab:ablation 1}
\end{table}




\subsection{Ablation Studies}
To prove the effectiveness of our proposed method, we conduct a series of experiments on S3DIS \cite{armeni2017joint}, evaluate on Area 5. Our baseline employs kernel point convolution with 15 kernel points. As shown in Table \ref{tab:ablation 1}, each time we add or replace a module while keeping the rest unchanged. 
First, combining feature propagation(FP) with PDS produces a $+2.3\%$ boost. An explanation is that PDS preserves shape patterns in every downsampling level, FP correctly retrieves the shape patterns by interpolating features from neighborhood at upsampling level. Grid sampling loses geometric information, and this incorrectness accumulates through multiple downsampling layers. 
Moreover, adding local feature attention module produces $+2.9\%$ gain.
Finally, SPConv improves the performance by $+3.4\%$, showing its great capability of local feature encoding.
Our full pipeline exceeds baseline with grid sample by $+4.5\%$, which has the state-of-the-art performance on S3DIS dataset \cite{armeni2017joint}.

To illustrate the effectiveness of our proposed attention module, see Table \ref{tab:attention}. Learnable MLP-based method achieves better performance however it burdens the computation.
\begin{table}[hbt!]
\caption{Ablation studies with the proposed feature attention.}
    \centering
    \begin{tabular}{C{9em} C{9em} C{6em} C{12em}}
    \hline
     Approach & Input features & mIoU & Inference speed (iter/s)\\
    \hline
    Gaussian function & color &  68.0 & 4.3\\
    Gaussian function & normal &  67.1 & 4.3\\
    Gaussian function & color+normal &  68.3 & 4.3\\
    \hline
    2layer MLP & color & 68.7 & 4.1\\
    2layer MLP & normal & 68.2 & 4.1\\
    2layer MLP & color+normal & 68.7 & 3.0\\
    3layer MLP & color+normal & 69.9 & 3.0\\
    \hline
    \end{tabular}
    \label{tab:attention}
\end{table}

\section{Conclusions}
In this work, we propose an architecture named SPNet for 3D point cloud semantic segmentation. We introduce a SPConv operator to effectively learn point cloud geometry. We demonstrate that with Poisson disk sampling as well as feature propagation, our network can go deep without losing much inherent shape patterns. Our framework outperforms many competing approaches proved by experimental results on public large-scale datasets. We will experiment our method on outdoor Lidar datasets and investigate more effective attention methods.

\bibliographystyle{splncs04}
\bibliography{egbib}
%




\end{document}